\documentclass{article}

\usepackage{arxiv}

\usepackage[utf8]{inputenc} 
\usepackage[T1]{fontenc}    
\PassOptionsToPackage{hyphens}{url}\usepackage{hyperref}     
\usepackage{url}            
\usepackage{booktabs}       
\usepackage{amsfonts}       
\usepackage{nicefrac}       
\usepackage{microtype}      
\usepackage{natbib}
\usepackage{doi}

\usepackage{makecell} 
\usepackage[export]{adjustbox}
\usepackage{subcaption}
\usepackage{placeins}
\usepackage{mathtools}
\usepackage[dvipsnames]{xcolor}
\usepackage{amsthm}

\newtheorem{theorem}{Theorem}

\newtheorem{definition}{Definition}

\newtheorem{assumption}{Assumption}

\DeclareMathOperator*{\argmax}{\arg\max}

\usepackage{cleveref}       

\title{Safe Policy Improvement Approaches\\ on Discrete Markov Decision Processes}

\author{Philipp Scholl \\
	Department of Mathematics\\
	Ludwig-Maximilians University\\
	Munich, Germany \\
	\texttt{scholl@math.lmu.de} \\
	\And
	Felix Dietrich \\
	Department of Computer Science\\
	Technical University of Munich\\
	Munich, Germany\\
	\texttt{felix.dietrich@tum.de} \\
	\And
	Clemens Otte \\
	Learning Systems \\
	Siemens Technology \\
	Munich, Germany \\
	\texttt{clemens.otte@siemens.com} \\
	\And
    Steffen Udluft \\
	Learning Systems \\
	Siemens Technology \\
	Munich, Germany \\
	\texttt{steffen.udluft@siemens.com} \\
}

\renewcommand{\shorttitle}{Safe Policy Improvement Approaches on Discrete Markov Decision Processes}

\hypersetup{
pdftitle={Safe Policy Improvement Approaches on Discrete Markov Decision Processes},
pdfauthor={Philipp~scholl, Felix~Dietrich, Clemens~Otte, Steffen~Udluft},
pdfkeywords={Risk-Sensitive Reinforcement Learning, Safe Policy Improvement, Markov Decision Processes},
}

\begin{document}
\maketitle

\begin{abstract}
	Safe Policy Improvement (SPI) aims at provable guarantees that a learned policy is at least approximately as good as a given baseline policy. Building on SPI with Soft Baseline Bootstrapping (Soft-SPIBB) by Nadjahi et al., we identify theoretical issues in their approach, provide a corrected theory, and derive a new algorithm that is provably safe on finite Markov Decision Processes (MDP). Additionally, we provide a heuristic algorithm that exhibits the best performance among many state of the art SPI algorithms on two different benchmarks. Furthermore, we introduce a taxonomy of SPI algorithms and empirically show an interesting property of  two classes of SPI algorithms: while the mean performance of algorithms that incorporate the uncertainty as a penalty on the action-value is higher, actively restricting the set of policies more consistently produces good policies and is, thus, safer.
\end{abstract}


\section{Introduction} \label{sec:introduction}
Reinforcement learning (RL) in industrial control applications such as gas turbine control \citep{schaefer_neural_2007} often requires learning a control policy solely on pre-recorded observation data, known as batch or offline RL \citep{lange_batch_2012,pmlr-v97-fujimoto19a,DBLP:journals/corr/abs-2005-01643}.
This is necessary because an online exploration on the real system or its simulation is not possible. 
Assessing the true quality of the learned policy is difficult in this setting \citep{5967358,wang2021what}. Thus, Safe Policy Improvement \citep{thomas_safe_nodate,nadjahi_safe_2019} is an attractive resort as it aims at ensuring that the learned policy is, with a high probability, at least approximately as good as a baseline policy given by, e.g., a conventional controller.

Safety is an overloaded term in Reinforcement Learning as it can refer to the inherent uncertainty, safe exploration techniques or parameter uncertainty
\citep{garcia_comprehensive_nodate}. In this paper we focus on the latter.

\subsection{Related work} \label{sec:related-work}

\begin{figure*}[h!]
  \centering
  \includegraphics[width = 1\linewidth]{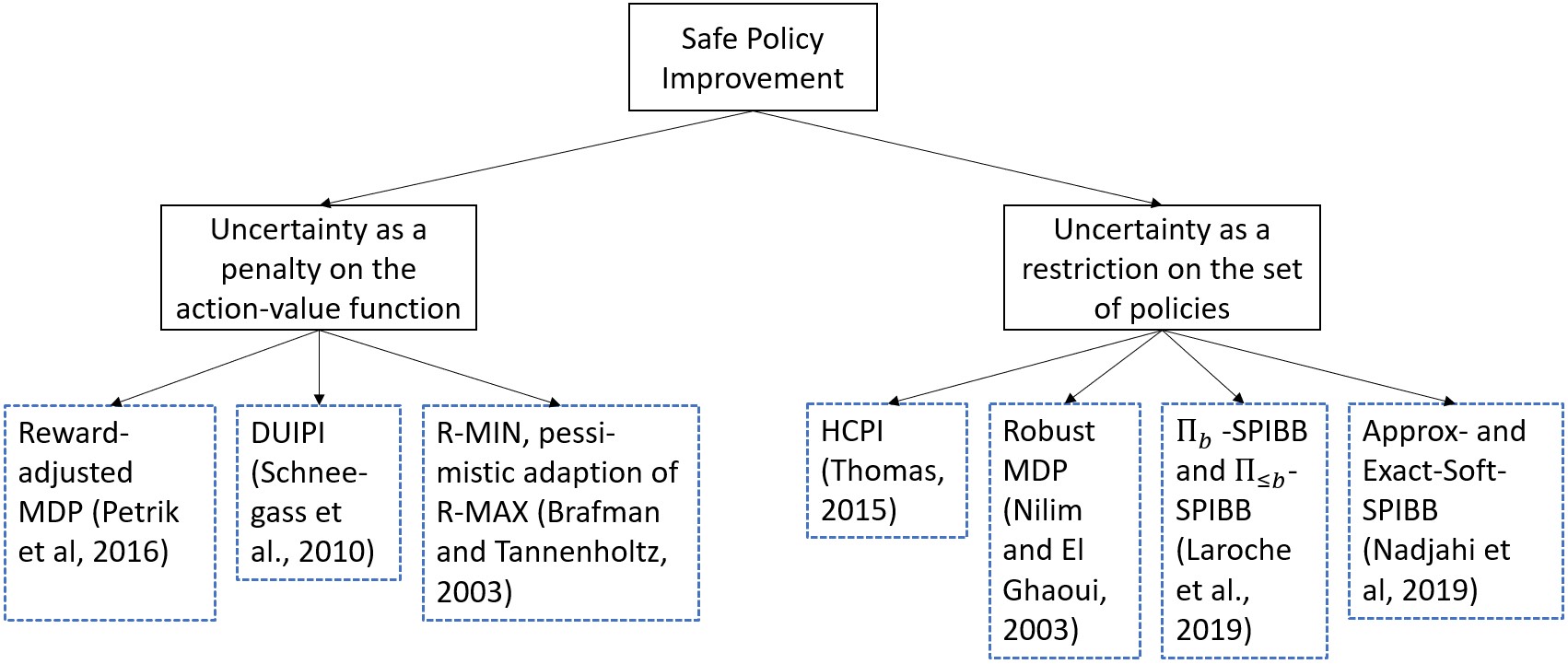}
  \caption{Taxonomy of SPI algorithms.}
  \label{fig:taxonomy}
 \end{figure*}

Many of the existing Safe Policy Improvement (SPI) algorithms utilize the uncertainty of state-action pairs in one of the two following ways (see also Figure \ref{fig:taxonomy}):
\begin{itemize}
    \item[1.] The uncertainty is applied to the action-value function to decrease the value of uncertain actions.
    \item[2.] The uncertainty is used to restrict the set of policies that can be learned.
\end{itemize}

\cite{thomas_safe_nodate} introduced High Confidence Policy Improvement (HCPI), an algorithm utilizing concentration inequalities on the importance sampling estimate of the performance of learned policies to ensure that the new policy is better than the baseline with a high probability. As HCPI simply rejects policies where the confidence intervals give no certain improvement, they essentially restrict the set of possible policies. This restriction is clearer for Robust MDP \citep{robust_mdps}, which computes the policy with the best worst-case performance for all transition probabilities in a convex set, which is chosen such that the true transition probabilities are part of it with a high probability. 

\cite{petrik} showed that maximizing the difference between the new policy and a baseline policy on a rectangular uncertainty set of the transition probabilities is NP-hard and, thus, derived the approximation Reward-adjusted MDP (RaMDP). RaMDP applies the uncertainty to penalize the reward and, therefore, the action-value function. \cite{laroche_safe_2019} extended this algorithm by a hyper-parameter which controls the influence of the uncertainty. RaMDP computes the uncertainty simply as a function of the number of visits to a state-action pair. A more sophisticated approach to estimate the uncertainty is taken for Diagonal Approximation of Uncertainty Incorporating Policy Iteration (DUIPI) in \cite{jabin_uncertainty_2010}, which estimates the standard deviation of the action-value function and applies this as a penalty to the action-value function.
Utilizing the uncertainty as an incentive instead of a penalty results in an explorative algorithm. 
Applying this correspondence between exploratory and safe behavior to further algorithms, one can easily adapt the efficiently exploring R-MAX \citep{Brafman2002RMAXA}, which assigns the highest value possible to all rarely visited state-action pairs, to its risk averse counterpart that we denote as R-MIN. This algorithm simply sets the action-value to the lowest possible value instead of the highest one for rarely visited state-action pairs.

Returning back to algorithms restricting the policy set, \cite{laroche_safe_2019} only allow deviations from the baseline policy at a state-action pair if the uncertainty is low, otherwise it remains the same. They propose two algorithms: $\Pi_b$-SPIBB, which is provably safe, and $\Pi_{\leq b}$-SPIBB, which is a heuristic relaxation. Both are tested against HCPI, Robust MDP, and RaMDP. HCPI and Robust MDP are strongly outperformed by the others and, while the mean performance of RaMDP is very good, its safety is inferior to both SPIBB algorithms. \cite{nadjahi_safe_2019} continue this line of work and relax the hard bootstrapping to a softer version, where the baseline policy can be changed at any state-action pair, but the amount of possible change is limited by the uncertainty at this state-action pair. They claim that these new algorithms, called Safe Policy Improvement with Soft Baseline Bootstrapping (Soft-SPIBB), are also provably safe, a claim that is repeated in \cite{Simo2020SafePI} and \cite{leurent:tel-03035705}. Furthermore, they extend the experiments from \cite{laroche_safe_2019} to include the Soft-SPIBB algorithms, where the empirical advantage of these algorithms becomes clear.  

\subsection{Our contributions}

We investigate the class of Soft-SPIBB algorithms \citep{nadjahi_safe_2019} and show that they are not provably safe. Hence, we derive the adaptation Adv-Approx-Soft-SPIBB which is provably safe. We also develop the heuristic Lower-Approx-Soft-SPIBB, following an idea presented in \cite{laroche_safe_2019}. Additionally, we conduct experiments to test these new versions against their predecessors and add further uncertainty incorporating algorithms \citep{Brafman2002RMAXA,jabin_uncertainty_2010} which were not considered in \cite{laroche_safe_2019} and \cite{nadjahi_safe_2019}. Here, we also show how the taxonomy illustrated in Figure \ref{fig:taxonomy} proves to be helpful, as both classes of algorithms present different behavior.
The code for the algorithms and experiments can be found in the accompanying repository.\footnote{\url{https://github.com/Philipp238/Safe-Policy-Improvement-Approaches-on-Discrete-Markov-Decision-Processes}}
\subsection{Outline}

The next section introduces the mathematical framework necessary for the later sections. Section \ref{sec:soft-spibb} begins with the work done by \cite{nadjahi_safe_2019} and ends with a discussion and proof of the shortcomings of the given safety guarantees. In Section \ref{sec:algorithms} we deduce the new algorithms, which will be tested against various competitors on two benchmarks in Section \ref{sec:experiments}.

\section{Mathematical framework}

The control problem we want to tackle with reinforcement learning consists of an agent and an environment, modeled as a finite Markov Decision Process (MDP). A finite MDP $M^*$ is represented by the tuple $M^*=(\mathcal{S}, \mathcal{A}, P^*, R^*, \gamma)$, where $\mathcal{S}$ is the finite state space, $\mathcal{A}$ the finite action space, $P^*$ the unknown transition probabilities, $R^*$ the unknown stochastic reward function, the absolute value of which is assumed to be bounded by $R_{max}$, and $0\leq\gamma<1$ is the discount factor.

The agent chooses action $a\in\mathcal{A}$ with probability $\pi(a|s)$ in state $s\in\mathcal{S}$, where $\pi$ is the policy controlling the agent. The return at time $t$ is defined as the discounted sum of rewards $G_t=\sum_{i=t}^{T}\gamma ^{i-t} R^*(s_i,a_i)$, with $T$ the time of termination of the MDP. As the reward function is bounded the return is bounded as well, since $|G_t|\leq\frac{R_{max}}{1-\gamma}$. So, let $G_{max}$ be a bound on the absolute value of the return. The goal is to find a policy $\pi$ which optimizes the expected return, i.e., the state-value function $V_{M^*}^{\pi}(s)=E_\pi[G_t|S_t=s]$ for the initial state $s\in\mathcal{S}$. Similarly, the action-value function is defined as $Q_{M^*}^{\pi}(s,a)=E_\pi[G_t|S_t=s,A_t=a]$.

Given data $\mathcal{D}=(s_j,a_j,r_j,s_j')_{j=1,\dots,n}$ collected by the baseline policy $\pi_b$, let $N_\mathcal{D}(s,a)$ denote the number of visits of the state-action pair $(s,a)$ in $\mathcal{D}$ and $\hat{M}=(\mathcal{S},\mathcal{A},\hat{P},\hat{R},\gamma)$ the Maximum Likelihood Estimator (MLE) of $M^*$ where
\begin{equation}
	\hat{P}(s'|s,a)=\frac{\sum_{(s_j=s,a_j=a,r_j,s_j'=s')\in\mathcal{D}}1}{N_\mathcal{D}(s,a)} \text{ and }   \hat{R}(s,a)=\frac{\sum_{(s_j=s,a_j=a,r_j,s_j')\in\mathcal{D}}r_j}{N_\mathcal{D}(s,a)}.
\end{equation}

\section{The Soft-SPIBB paradigm}
 \label{sec:soft-spibb}

The idea in \cite{nadjahi_safe_2019} is to estimate the uncertainty in the state-action pairs and bound the change in the baseline policy accordingly.

\subsection{Preliminaries}

To bound the performance of the new policy it is necessary to bound the estimate of the action-value function. In \cite{nadjahi_safe_2019} this is done for $Q_{\hat{M}}^{\pi}$ by applying Hoeffding's inequality. However, Hoeffding's inequality is only applicable for the arithmetic mean of independent, bounded random variables, thus, we define $\hat{Q}^{\pi_b}_{\mathcal{D}}(s,a)=\frac{1}{n}\sum_{i=1}^nG_{t_i}$ as the Monte Carlo estimate of the action-value function, where $t_1,...,t_n$ are times such that $(S_{t_i},A_{t_i})=(s,a)$ for all $i=1,...,n$. See \cite{scholl} for a discussion of the (approximate) independence of $G_{t_i}$. Following the proof in Appendix A.2 in \cite{nadjahi_safe_2019} yields that
\begin{equation} \label{eq:bound-q}
	|Q_{M^*}^{\pi_b}(s,a) - \hat{Q}^{\pi_b}_{\mathcal{D}}(s,a)|\leq e_Q(s,a)G_{max}
\end{equation}
holds with probability $1-\delta$ for all state-action pairs. Here, $e_Q$ is the error function computing the uncertainty of one state-action pair and is given by
\begin{equation} \label{eq:error-function}
	e_Q(s,a)=\sqrt{\frac{2}{N_\mathcal{D}(s,a)}\log\frac{2|\mathcal{S}||\mathcal{A}|}{\delta}}.
\end{equation} 

Analogously, 
\begin{equation} \label{eq:bound-P}
	||P(\cdot|s,a)-\hat{P}(\cdot|s,a)||_1\leq e_P(s,a),
\end{equation}
holds with probability $1-\delta$ where 
\begin{equation} \label{eq:error-function-P}		
	e_P(s,a)=\sqrt{\frac{2}{N_\mathcal{D}(s,a)}\log\frac{2|\mathcal{S}||\mathcal{A}|2^{|\mathcal{A}|}}{\delta}}.
\end{equation}

The error functions are used to quantify the uncertainty of each state-action pair.
\begin{definition} \label{def:constrained}
	A policy $\pi$ is $(\pi_b,\epsilon,e)$-\emph{constrained} w.r.t.\ a baseline policy
	$\pi_b$, an error function $e$ and a hyper-parameter $\epsilon>0$, if 
	\begin{equation} \label{eq:constrained}
		\sum_{a\in\mathcal{A}}e(s,a)|\pi(a|s)-\pi_b(a|s)|\leq\epsilon
	\end{equation}
	holds for all states $s\in\mathcal{S}$.
\end{definition}
So, if a policy $\pi$ is $(\pi_b,\epsilon,e)$-constrained, it means that the $l^1$-distance between $\pi$ and $\pi'$, weighted by some error function $e$, is at most $\epsilon$. To utilize Equation \ref{eq:bound-q} later the following property is also necessary:	\begin{definition} \label{def:advantageous-wrt-function}
	A policy $\pi$ is $\pi_b$-\emph{advantageous w.r.t.\ the function}
	$Q:\mathcal{S}\times\mathcal{A}\rightarrow\mathbb{R}$, if 
	\begin{equation} \label{eq:def-advantageous-wrt-function}
		\sum_a Q(s,a)\pi(a|s)\geq\sum_a Q(s,a)\pi_b(a|s)
	\end{equation}
	holds for all states $s\in\mathcal{S}$. 
\end{definition}
Note that this is an extension of Definition 3 in \cite{nadjahi_safe_2019} to arbitrary functions and can, thus, be used for $\hat{Q}^{\pi_b}_{\mathcal{D}}$. Interpreting $Q$ as some kind of action-value function, Definition \ref{def:advantageous-wrt-function} gives that the policy $\pi$ chooses higher valued actions than policy $\pi'$ for every state.

\subsection{The Algorithms} \label{sec:soft-spibb-algorithms}

The new class of algorithms \citep{nadjahi_safe_2019} introduce make use of the classical Policy Evaluation and Policy Improvement scheme \citep{sutton_reinforcement_nodate}, where the Policy Evaluation step is completely analogous to the one for dynamic programming with estimated model parameters $\hat{P}$ and $\hat{R}$. The Policy Improvement step, however, aims at solving the constrained optimization problem:
\begin{equation} \label{eq:optimization-problem-PI}
	\pi^{(i+1)}=\argmax_\pi\sum_{a\in\mathcal{A}}Q^{\pi^{(i)}}_{\hat{M}}(s,a)\pi(a|s)
\end{equation}
subject to:\\
\textbf{Constraint 1:} $\pi^{(i+1)}(\cdot|s)$ being a probability over
$\mathcal{A}$: $\sum_{a\in\mathcal{A}}\pi^{(i+1)}(a|s)=1$ and $\forall
a\in\mathcal{A}:\pi^{(i+1)}(a|s)\geq0$.\\
\textbf{Constraint 2:} $\pi^{(i+1)}$ being $(\pi_b,\epsilon, e)$-constrained.\\

Thus, it tries to compute the optimal---w.r.t. the action-value function of the previous policy---$(\pi_b,\epsilon, e)$-constrained policy. The two algorithms introduced in \cite{nadjahi_safe_2019} solving this optimization problems are Exact-Soft-SPIBB and Approx-Soft-SPIBB. The former solves the linear formulation of the constrained problem by a linear program \citep{linear-program} and the latter uses a budget calculation for Constraint 2 to compute an approximate solution. In experiments, it is shown that both algorithms achieve similar performances, but Exact-Soft-SPIBB takes considerably more time \citep{nadjahi_safe_2019}.

\subsection{The Safety Guarantees}

\cite{nadjahi_safe_2019} derive the theoretical safety of their algorithms from the following two theorems. Theorem \ref{th:theorem-1} shows that the performance of a policy which fulfills the two properties from Definitions \ref{def:constrained} and \ref{def:advantageous-wrt-function}, where the error function $e$ is such that Equation \ref{eq:bound-q} holds, can be bounded from below with a high probability.

\begin{theorem}\label{th:theorem-1}
	For any $(\pi_b,\epsilon,e_Q)$-constrained policy that is $\pi_b$-advantageous
	w.r.t.\ $\hat{Q}^{\pi_b}_{\mathcal{D}}$, which is estimated with independent
	returns for each state-action pair, the following inequality holds:
	\begin{equation}
		\mathbb{P}_\mathcal{D}\left(\forall
		s\in\mathcal{S}:V_{M^*}^{\pi}(s)-V_{M^*}^{\pi_b}(s)\geq-\frac{\epsilon
			G_{max}}{1-\gamma}\right)\geq 1-\delta,\nonumber
	\end{equation}
	where $M^*$ is the true MDP on which the data $\mathcal{D}$ gets sampled by the
	baseline policy $\pi_b$, $0\leq\gamma<1$ is the discount factor, and $\delta>0$
	is the safety parameter for $e_Q$.
\end{theorem}

This is essentially the same as Theorem 1 in \cite{nadjahi_safe_2019} if Equation \ref{eq:bound-q} from this paper is used instead of Equation 2 in \cite{nadjahi_safe_2019}. A full version of the proof with an accompanying thorough discussion can be found in \cite{scholl}. 

The  optimization problem solved by the Soft-SPIBB algorithms, however, does not enforce that the new policy is $\pi_b$-advantageous w.r.t.\ $\hat{Q}^{\pi_b}_{\mathcal{D}}$ and, so, Theorem 1 cannot be applied to them. Therefore, \cite{nadjahi_safe_2019} prove Theorem \ref{th:theorem-2} by assuming the following:

\begin{assumption} \label{ass:assumption-1}
    There exists a constant $\kappa<\frac{1}{\gamma}$ such
	that, for all state-action pairs $(s,a)\in\mathcal{S}\times\mathcal{A}$, the
	following holds:
    \begin{equation} \label{eq:assumption-1}
	    \sum_{s',a'}e_P(s',a')\pi_b(a'|s')P^*(s'|s,a)\leq\kappa e_P(s,a)
    \end{equation}
\end{assumption}

Interpreting $\pi_b(a'|s')P^*(s'|s,a)$ as the probability of observing the state-action pair $(s',a')$ after observing $(s,a)$ we can rewrite Equation \ref{eq:assumption-1} to 
\begin{equation} \label{eq:assumption-1-reformulated}
	E_{P,\pi_b}[e_P(S_{t+1},A_{t+1})|S_t=s,A_t=a]\leq\kappa e_P(s,a)
\end{equation}
which shows that Assumption \ref{ass:assumption-1} assumes an upper bound on the expected number of visits of the next state-action pair dependent on the number of visits of the current one. This might intuitively make sense, but we show in the next section that it is wrong in general. However, using this assumption \cite{nadjahi_safe_2019} prove Theorem \ref{th:theorem-2} which omits the advantageous assumption of the new policy.

\begin{theorem} \label{th:theorem-2}
    Under Assumption 1, any
		$(\pi_b,\epsilon,e_P)$-constrained policy $\pi$ satisfies the following
		inequality in every state $s$ with probability at least $1-\delta$:\\
	\begin{gather}
		V^\pi_{M^*}(s)-V^{\pi_b}_{M^*}(s)\geq V^\pi_{\hat{M}}(s)-V^{\pi_b}_{\hat{M}}(s) +\nonumber\\
		2||d_\pi^M(\cdot|s)-d_{\pi_b}^M(\cdot|s)||_1v_{max}
		-\frac{1+\gamma}{(1-\gamma)^2(1-\kappa\gamma)}\epsilon v_{max}
	\end{gather}
\end{theorem}
Here, $d_\pi^M(s'|s)$ denotes the expected discounted sum of visits to $s'$ when starting in $s$. 

\subsection{Shortcomings of the Theory}
As explained above, the theoretical guarantees, \cite{nadjahi_safe_2019} claim for the Soft-SPIBB algorithms, stem from Theorem \ref{th:theorem-1} and \ref{th:theorem-2}. However, Theorem \ref{th:theorem-1} is only applicable to policies which are $\pi_b$-advantageous w.r.t. $\hat{Q}^{\pi_b}_{\mathcal{D}}$ and Theorem \ref{th:theorem-2} relies on Assumption \ref{ass:assumption-1}. In the following we show in Theorem \ref{th:assumption-1} that Assumption \ref{ass:assumption-1} does not hold for any $0<\gamma<1$.

\begin{theorem}	\label{th:assumption-1}
	Let the discount factor $0<\gamma<1$ be arbitrary. Then there exists an MDP $M$
	with transition probabilities $P$ such that for any behavior policy $\pi_b$ and
	any data set $\mathcal{D}$, which contains every state-action pair at least once, it holds that, for all $0<\delta<1$,
	\begin{equation} \label{eq:assumption-1-theorem}
		\sum_{s',a'}e_P(s',a')\pi_b(a'|s')P(s'|s,a) > \frac{1}{\gamma} e_P(s,a).
	\end{equation}
	This means that Assumption 1 can, independent of the discount factor, not be
	true for all MDPs.
\end{theorem}
\begin{figure}[!h]
  \centering
  \includegraphics[width = 4.5cm]{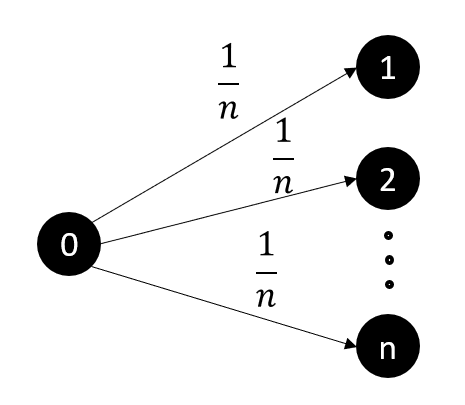}
  \caption{MDP with $n+1$ states, $n$ of them are final states and in the non-final state, there is only $1$ action, leading to one of the others with equal probability.}
  \label{fig:example-mdp-assumption-1}
\end{figure}
\begin{proof}
 	Let $0<\gamma<1$ be arbitrary and $n\in\mathbb{N}$ be such that
	$\sqrt{n}>\frac{1}{\gamma}$.
		Let $M$ be the MDP displayed in Figure \ref{fig:example-mdp-assumption-1}. It
	has $n+1$ states, from which $n$ states are terminal states, labeled $1$, $2$,
	..., $n$. In the only non-terminal state $0$, there is only one action available
	and choosing it results in any of the terminal states with probability
	$\frac{1}{n}$. As there is only one
	action, one can omit the action in the notation of $e_P$ and there is only one possible behavior policy. So, Equation
	\ref{eq:assumption-1-theorem} can be reduced to
	\begin{equation} \label{eq:assumption-1-proof-one-action}
		\sum_{i=1}^n\frac{e_P(i)}{n} > \frac{e_P(0)}{\gamma} .
	\end{equation}
	Now, we show that 
	\begin{equation} \label{eq:assumption-1-proof-goal}
		\sum_{i=1}^n\frac{e_P(i)}{n} \geq \sqrt{n}e_P(0),
	\end{equation}
	which implies Equation
	\ref{eq:assumption-1-proof-one-action} as
	$\sqrt{n}>\frac{1}{\gamma}$. Let $\mathcal{D}$ denote the data collected on this MDP such that every state has been visited at least once. Thus, $N_\mathcal{D}(i)>0$\textemdash{}the number of visits to state $i$\textemdash{}holds for every $i$. Equation \ref{eq:assumption-1-proof-goal} is equivalent to 
	\begin{equation} \label{eq:assumption-1-proof-equivalent-goal}
		\frac{1}{n}\sum_{i=1}^n\frac{1}{\sqrt{N_\mathcal{D}(i)}} \geq
		\frac{\sqrt{n}}{\sqrt{N}}
	\end{equation}
	where $N=N_\mathcal{D}(0)=\sum_{i=1}^nN_\mathcal{D}(i)$.  Equation \ref{eq:assumption-1-proof-equivalent-goal} follows by applying Jensen's inequality once for the convex function $x\mapsto\frac{1}{x}$, restricted to $x>0$, and once for the concave function $x\mapsto\sqrt{x}$, also restricted to $x>0$:
	\begin{gather} \label{eq:end-proof-ass-1}
	    \frac{1}{n}\sum_{i=1}^n\frac{1}{\sqrt{N_\mathcal{D}(i)}} \geq 
	    \frac{1}{\frac{1}{n}\sum_{i=1}^n\sqrt{N_\mathcal{D}(i)}} \geq \nonumber
	    \frac{1}{\sqrt{\frac{1}{n}\sum_{i=1}^n N_\mathcal{D}(i)}} = 
	    \frac{1}{\sqrt{\frac{N}{n}}} = 
	    \frac{\sqrt{n}}{\sqrt{N}}.
	\end{gather}
\end{proof}
	The class of MDPs used in the proof and depicted in Figure
	\ref{fig:example-mdp-assumption-1} gives a good impression what kind of
	constellations are critical for Assumption 1. An MDP does not have to exhibit exactly
	the same structure to have similar effects, it might already be enough if there is
	a state-action pair from which a lot of different states-action pairs are
	exclusively accessible. 
	
	A reasonable question is whether although Assumption 1 is invalid in its
	generality shown at some specific class of MDPs it might hold on simple
	MDPs which are not built in order to disprove Assumption 1. One consideration
	here is that $n$ does not need to be especially big as the proof only required
	$\sqrt{n}>\frac{1}{\gamma}$. So, for any $\gamma>\frac{1}{\sqrt{2}}\approx
	0.707$ it suffices to choose $n=2$. 
	
    Furthermore, we tested Assumption \ref{ass:assumption-1} empirically on the Random MDPs benchmark considered in \cite{nadjahi_safe_2019} where we found for no discount factor greater than 0.6 a baseline policy and data set such that the assumption holds for all state-action pairs.\footnote{\url{https://github.com/Philipp238/Safe-Policy-Improvement-Approaches-on-Discrete-Markov-Decision-Processes/blob/master/auxiliary\_tests/assumption\_test.py}}
    
    Consequently, we conclude that Assumption \ref{ass:assumption-1} is not reasonable and, thus, Theorem \ref{th:theorem-2} cannot be relied upon. As mentioned before, Theorem \ref{th:theorem-1} is only applicable to $\pi_b$-advantageous w.r.t. $\hat{Q}^{\pi_b}_{\mathcal{D}}$ policies. For this reason, both Soft-SPIBB algorithms are not provably safe.

\section{Algorithms}\label{sec:algorithms}
In this section we introduce the adaptation Adv-Approx-Soft-SPIBB which produces ($\pi_b,e_Q,\epsilon)$-constrained and $\pi_b$-advantageous w.r.t. $\hat{Q}^{\pi_b}_{\mathcal{D}}$ policies and, thus, Theorem \ref{th:theorem-1} is applicable to it, making it provably safe. Additionally, we present the heuristic adaptation Lower-Approx-Soft-SPIBB. As both algorithms function similarly to their predecessors by constraining the policy set, they also belong to the category "Uncertainty as a restriction on the set of policies" in the taxonomy in Figure \ref{fig:taxonomy}.

\subsection{Adv-Approx-Soft-SPIBB}
The advantageous version of the Soft-SPIBB algorithms solve the following optimization problem in the Policy Improvement (PI) step:
\begin{equation} \label{eq:optimization-problem-PI-adv}
		\pi^{(i+1)}=\argmax_\pi\sum_{a\in\mathcal{A}}Q^{\pi^{(i)}}_{\hat{M}}(s,a)\pi(a|s)
\end{equation}
subject to:\\
\textbf{Constraint 1:} $\pi^{(i+1)}(\cdot|s)$ being a probability over
$\mathcal{A}$: $\sum_{a\in\mathcal{A}}\pi^{(i+1)}(a|s)=1$ and $\forall
a\in\mathcal{A}:\pi^{(i+1)}(a|s)\geq0$.\\
\textbf{Constraint 2:} $\pi^{(i+1)}$ being $(\pi_b,\epsilon, e)$-constrained.\\
\textbf{Constraint 3:} $\pi^{(i+1)}$ being $\pi_b$-advantageous w.r.t.
$\hat{Q}^{\pi_b}_\mathcal{D}$.

The original Soft-SPIBB algorithms solve this optimization problem without Constraint 3 as shown in Section \ref{sec:soft-spibb-algorithms}. To solve the problem including constraint 3, we introduce Adv-Approx-Soft-SPIBB. This algorithm works exactly as its predecessor Approx-Soft-SPIBB except that it keeps an additional budgeting variable ensuring that the new policy is $\pi_b$-advantageous w.r.t. $\hat{Q}^{\pi_b}_\mathcal{D}$.

The derivation of a successor algorithm of Exact-Soft-SPIBB is straightforward since Constraint 3 is linear, however, we observed for Exact-Soft-SPIBB and its successor numerical issues, so, we omit them in the experiments in Section \ref{sec:experiments}.

\subsection{Lower-Approx-Soft-SPIBB}

To introduce the heuristic adaptation of Approx-Soft-SPIBB we need a relaxed version of the constrainedness property.

\begin{definition} \label{def:lower-constrained}
	A policy $\pi$ is $(\pi_b,\epsilon,e)$-\emph{lower-constrained} w.r.t.\ a baseline
	policy $\pi_b$, an error function $e$, and a hyper-parameter $\epsilon$, if 
	\begin{equation} \label{eq:lower-constrained}
		\sum_{a\in\mathcal{A}}e(s,a) \max\{0, \pi(a|s)-\pi_b(a|s)\}\leq\epsilon
	\end{equation}
	holds for all states $s\in\mathcal{S}$.
\end{definition}

This definition does not punish a change in uncertain state-action pairs if the probability of choosing it is decreased, which follows the same logic as the empirically very successful adaptation $\Pi_{\leq b}$-SPIBB \citep{laroche_safe_2019}. The optimization problem solved by Lower-Approx-Soft-SPIBB is the following:
\begin{equation} \label{eq:optimization-problem-PI-lower-approx-soft}
	\pi^{(i+1)}=\argmax_\pi\sum_{a\in\mathcal{A}}Q^{\pi^{(i)}}_{\hat{M}}(s,a)\pi(a|s)
\end{equation}
subject to:\\
\textbf{Constraint 1:} $\pi^{(i+1)}(\cdot|s)$ being a probability over
$\mathcal{A}$: $\sum_{a\in\mathcal{A}}\pi^{(i+1)}(a|s)=1$ and $\forall
a\in\mathcal{A}:\pi^{(i+1)}(a|s)\geq0$.\\
\textbf{Constraint 2:} $\pi^{(i+1)}$ being $(\pi_b,\epsilon,e)$-lower-con\-strained.

Even though Lower-Approx-Soft-SPIBB is\textemdash{}just as its predecessor Approx-Soft-SPIBB\textemdash{}not provably safe, the experiments in Section \ref{sec:experiments} show that it performs empirically the best out of the whole SPIBB family.

\section{Experiments}\label{sec:experiments}
We test the new Soft-SPIBB algorithms against Basic RL (classical Dynamic Programming \citep{sutton_reinforcement_nodate} on the MLE MDP $\hat{M}$, Approx-Soft-SPIBB \citep{nadjahi_safe_2019}, its predecessors,  $\Pi_b$- and $\Pi_{\leq b}$-SPIBB \citep{laroche_safe_2019}, DUIPI \citep{jabin_uncertainty_2010}, RaMDP \citep{petrik} and R-MIN, the pessimistic adaptation of R-MAX \citep{Brafman2002RMAXA}. We omit HCPI \citep{thomas_safe_nodate} and Robust MDPs \citep{robust_mdps} due to their inferior performance compared to the SPIBB and Soft-SPIBB algorithms reported in \cite{laroche_safe_2019} and \cite{nadjahi_safe_2019}.

We use two different benchmarks for our comparison. The first one is the Random MDPs benchmark already used in \cite{laroche_safe_2019} and \cite{nadjahi_safe_2019}. As the second benchmark we use the Wet Chicken benchmark \citep{alippi_efficient_2009} which depicts a more realistic scenario.

We perform a grid-search to choose the optimal hyper-parameter for each algorithm for both benchmarks. Our choices can be found in the table below.
\begin{table}[h]
\caption{Chosen hyper-parameters for both benchmarks.} \label{tab:hyper-parameter} \centering
\begin{tabular}{|c|c|c|}
	\hline
	\textbf{Algorithms} & \makecell{\textbf{Random} \\ \textbf{MDPs}} & \textbf{Wet Chicken} \\
	\hline
    Basic RL & - & - \\
	\hline
	RaMDP & $\kappa=0.05$ & $\kappa=2$ \\
	\hline
	R-MIN & $N_\wedge=3$ & $N_\wedge=3$ \\
	\hline
	DUIPI & $\xi=0.1$ & $\xi=0.5$ \\
	\hline
	$\Pi_b$-SPIBB & $N_\wedge=10$ & $N_\wedge=7$\\
	\hline
    $\Pi_{\leq b}$-SPIBB &$N_\wedge=10$ & $N_\wedge=7$\\
	\hline
	\makecell{Approx \\ -Soft-SPIBB} & $\delta=1,\,\epsilon=2$ & $\delta=1,\,\epsilon=1$\\
	\hline
	\makecell{Adv-Approx \\ -Soft-SPIBB \\ (ours)} & $\delta=1,\,\epsilon=2$ & $\delta=1,\,\epsilon=1$\\
	\hline
	\makecell{Lower-Approx \\ -Soft-SPIBB \\ (ours)} & $\delta=1,\,\epsilon=1$ & $\delta=1,\,\epsilon=0.5$\\

	\hline

\end{tabular}
\end{table}

\subsection{Random MDP Benchmark}
We consider the grid-world Random MDPs benchmark introduced in \cite{nadjahi_safe_2019} which generates a new MDP in each iteration. The generated MDPs consist of 50 states, including an initial state (denoted by 0) and a final state. In every non-terminal state there are four actions available and choosing one leads to four possible next states. All transitions yield zero reward except upon entering the terminal state, which gives a reward of 1. As the discount factor is chosen as $\gamma=0.95$, maximizing the return is equivalent to finding the shortest route to the terminal state. 

The baseline policy on each MDP is computed such that its performance is approximately $\rho_{\pi_b}=V_{M^*}^{\pi_b}(0)=\eta V_{M^*}^{\pi_*}(0)+(1-\eta)V_{M^*}^{\pi_u}(0)$, where $0\leq\eta\leq1$ is the baseline performance target ratio interpolating between the performance of the optimal policy $\pi_*$ and the uniform policy $\pi_u$. The generation of the baseline policy starts with a softmax on the optimal action-value function and continues with adding random noise to it, until the desired performance is achieved \citep{nadjahi_safe_2019}. To counter the effects from incorporating knowledge about the optimal policy, the MDP is altered after the generation of the baseline policy by transforming one regular state to a terminal one, called \emph{good easter egg}, which also yields a reward of 1.

In this experiment, $10,000$ iterations were run and the performances are
normalized to make them more comparable between different runs by calculating 
$\bar{\rho}_\pi= \frac{\rho_\pi-\rho_{\pi_b}}{\rho_{\pi_*}-\rho_{\pi_b}}$.
Thus, $\bar{\rho}_\pi<0$ means a worse performance than the baseline policy,
$\bar{\rho}_\pi>0$ means an improvement w.r.t.\ the baseline policy and
$\bar{\rho}_\pi=1$ means the optimal performance was reached. As we are interested in Safe Policy Improvement, we follow \cite{Chow2015RiskSensitiveAR}, \cite{laroche_safe_2019}, and \cite{nadjahi_safe_2019} and consider besides the mean performance also the 1\%-CVaR (Critical Value at Risk) performance, which is the mean performance over the 1\% worst runs.

\begin{figure*}[ht]
	\centering
	\begin{subfigure}{1\textwidth}
		\centering
		\includegraphics[width=\textwidth]{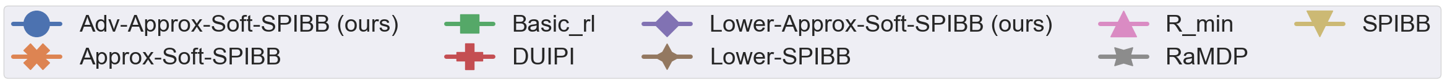}
		\vspace{0.2cm}
	\end{subfigure}
    	\hfill
	\begin{subfigure}[b]{0.35\textheight}
		\centering
		\includegraphics[width=\textwidth]{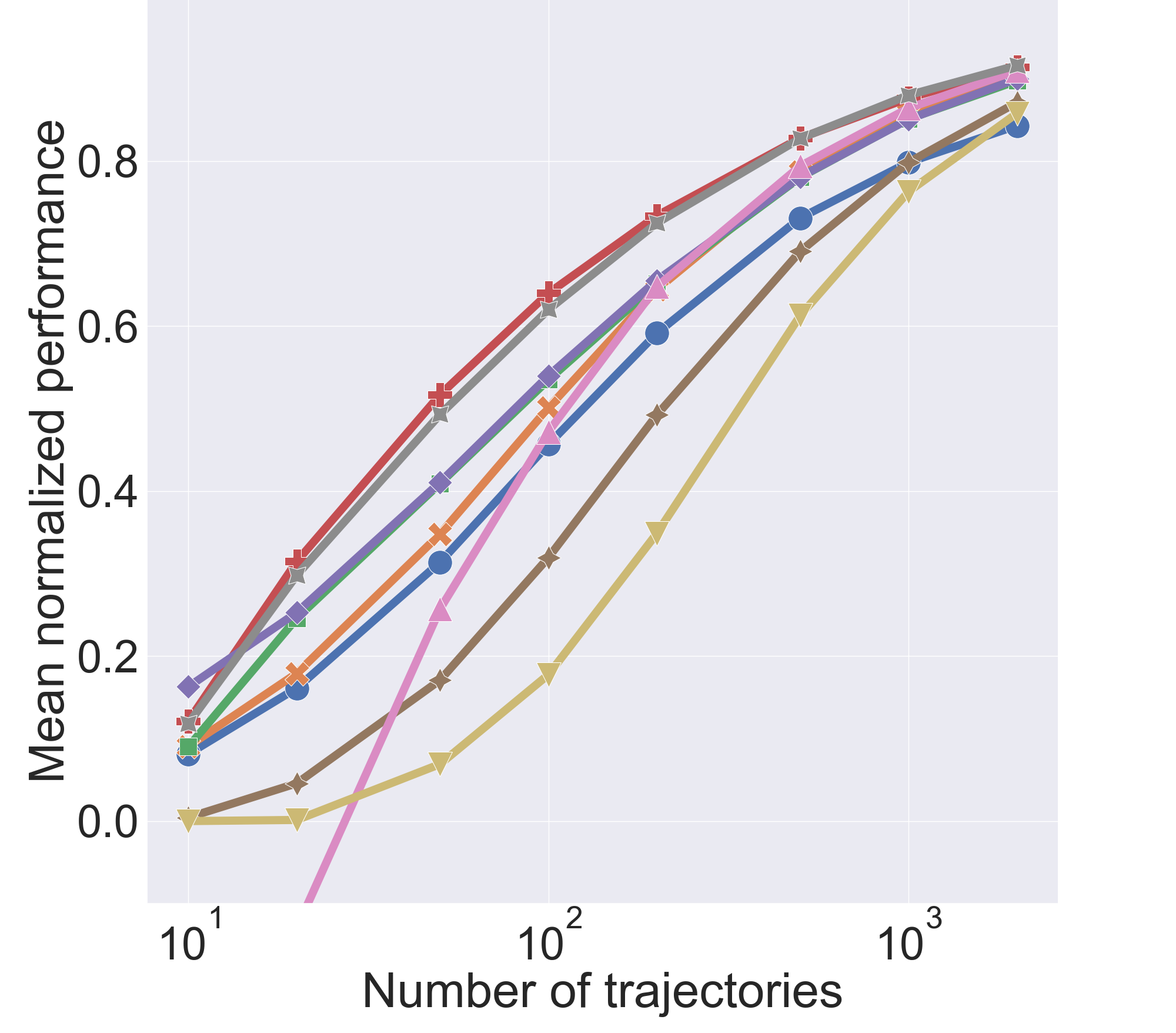}
		\caption{Mean}
		\label{fig:random-mdps-mean}
	\end{subfigure}
	\hfill
    \begin{subfigure}[b]{0.35\textheight}
		\centering
		\includegraphics[width=\textwidth]{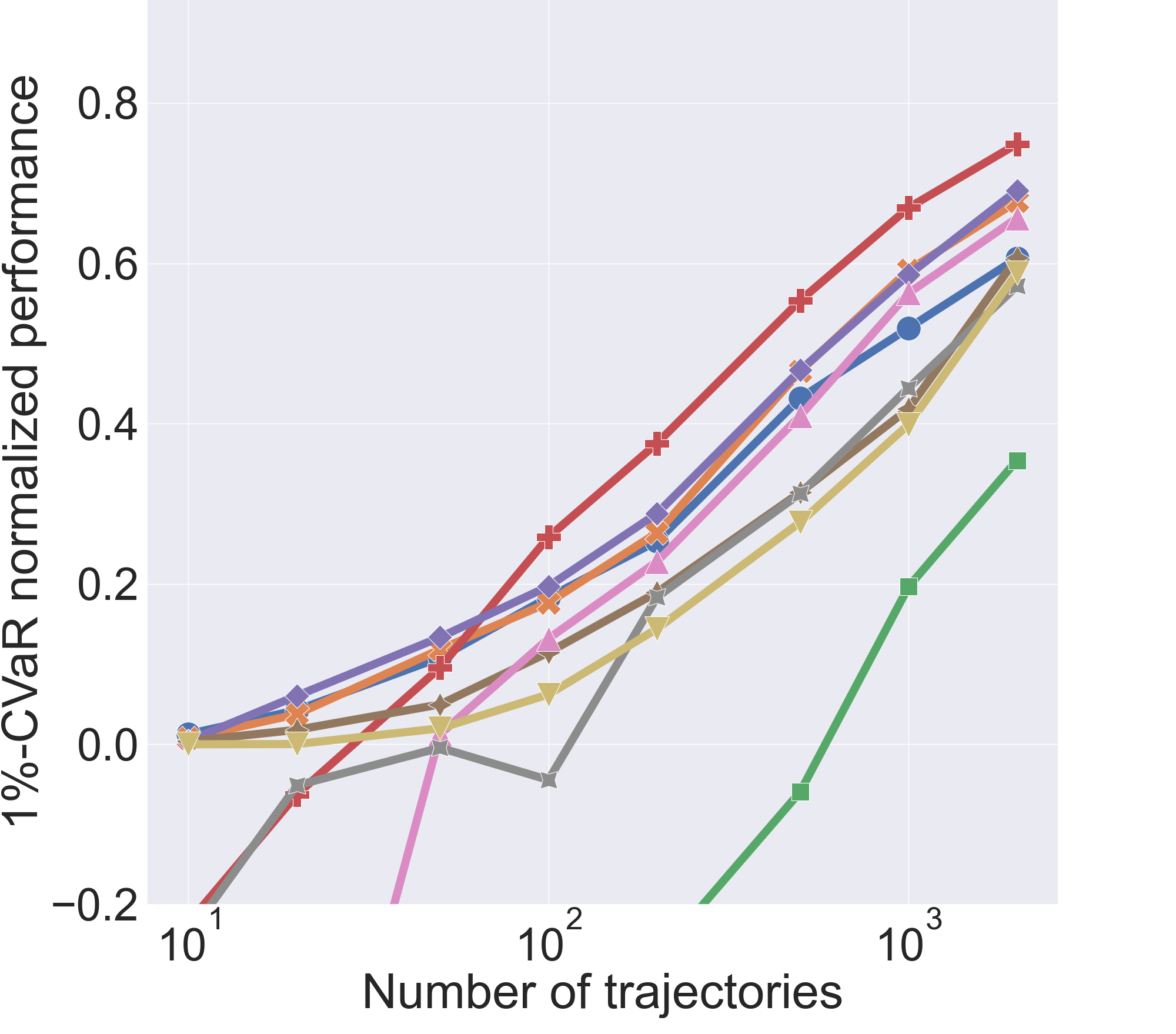}
		\caption{1\%-CVaR}
		\label{fig:random-mdps-cvar}
	\end{subfigure}
	\caption{Mean (a) and 1\%-CVaR (b) normalized performance over 10,000 trials on the Random MDPs benchmark for $\rho_{\pi_b}=0.9$. In the context of SPI the focus lies on the 1\%-CVaR. The mean performance is dominated by the algorithms applying a penalty on the action-value function, while the restricting algorithms are winning for few data points in the risk-sensitive 1\%-CVaR measure and only lose to DUIPI in the long run. Among the SPIBB class, Lower-Approx-Soft-SPIBB shows the best performance in both runs.}
	\label{fig:random-mdps}
\end{figure*}

These two measures can be seen in Figure \ref{fig:random-mdps}, where we show the performance of the algorithms for their optimal respective hyper-parameter, as displayed in Table \ref{tab:hyper-parameter}. In the mean performance, RaMDP and DUIPI outperform every other algorithm as soon as at least 20 trajectories are observed. They are followed by Lower-Approx-Soft-SPIBB and Basic RL, which come just before the other Soft-SPIBB algorithms. Approx-Soft-SPIBB shows a slightly better performance than its successor Adv-Approx-Soft-SPIBB. $\Pi_b$-SPIBB (SPIBB) and $\Pi_{\leq b}$-SPIBB (Lower-SPIBB) seem to perform the worst. While R-MIN exhibits issues when the data set is very small, it catches up with the others for bigger data sets. Interestingly, Basic RL performs generally quite well, but is still
outperformed by some algorithms, which might be surprising as the others are
intended for safe RL instead of an optimization of their mean performance. The
reason for this might be that considering the uncertainty of the action-value
function is even beneficial for the mean performance. 

The performance in the worst percentile looks very different. Here, it can be seen how
well the safety mechanisms of some of the algorithms work, particularly when compared to Basic RL,
which shows the worst overall 1\%-CVaR performance. Also, R-MIN and RaMDP
perform very poorly especially for a low number of trajectories. For less than 100 trajectories, DUIPI performs also very poorly but outperforms every other algorithm for bigger data sets. An interesting observation is that all the SPIBB
and Soft-SPIBB algorithms perform in the beginning extremely well, which is expected as they fall back to the behavior policy if not much
data is available. The ranking in the SPIBB family stays the same  for the 1\%-CVaR as it has been for the mean performance: Lower-Approx-Soft-SPIBB performs the best, closely followed first by Approx-Soft-SPIBB and Adv-Approx-Soft-SPIBB. $\Pi_{\leq b}$-SPIBB falls a bit behind but still manages to perform better than the original $\Pi_b$-SPIBB.

\subsection{Wet Chicken Benchmark}

	\begin{figure}[h!]
		\centering
		\includegraphics[width=5.9cm]{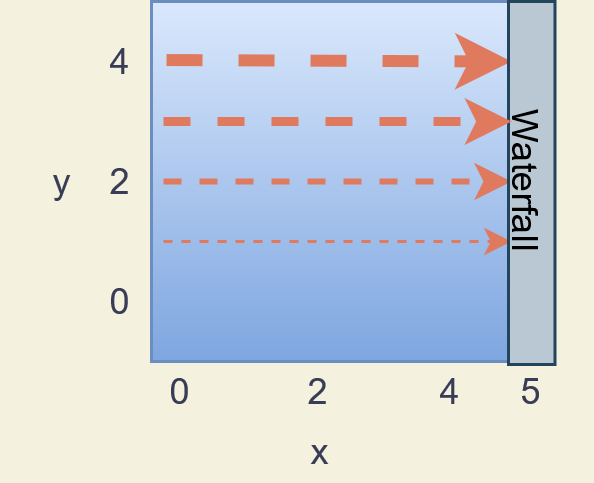}
		\caption{The setting of the Wet Chicken benchmark used for reinforcement learning. The boat starts at $(x,y)=(0,0)$ and starts there again upon falling down the waterfall at $x=5$. The arrows show the direction and strength of the stream towards the waterfall. Additionally, there are turbulences which are stronger for small $y$. The goal for the boat is to stay as close as possible to the waterfall without falling down.}
		\label{fig:wetchicken2}
	\end{figure}
	
Besides reproducing the results of \cite{nadjahi_safe_2019} for additional
	algorithms, we extend their experiments to a more realistic scenario for which we have chosen the discrete version of the Wet Chicken benchmark \citep{alippi_efficient_2009} because of its heterogeneous stochasticity. Figure \ref{fig:wetchicken2} visualizes the setting of the Wet Chicken benchmark. The basic idea behind it is that a person floats in a small boat on a river. The river has a waterfall at one end and the goal of the person is to stay as close to the waterfall as possible without falling down. Thus, the closer the person is to the waterfall the higher the reward gets, but upon falling down they start again at the starting place, which is as far away from the waterfall as possible. Therefore, this is modeled as a non-episodic MDP.
	
	The whole river has a length and width of 5, so, there are 25 states. The
	starting point is $(x,y)=(0,0)$ and the waterfall is at $x=5$. The position of
	the person at time $t$ is denoted by the pair $(x_t,y_t)$. The river itself has
	a turbulence which is stronger near the shore the person starts close to ($y=0$)
	and a stream towards the waterfall which is stronger near the other shore
	($y=4$). The velocity of the stream is defined as $v_t=y_t\frac{3}{5}$ and the
	turbulence as $b_t=3.5 - v_t$. The effect of the turbulence is stochastic; so, let
	$\tau_t\sim U(-1,1)$ be the parameter describing
	the stochasticity of the turbulence at time $t$.
	
	\begin{figure*}[ht]
	\centering
	\begin{subfigure}{1\textwidth}
		\centering
		\includegraphics[width=\textwidth]{Images/Legend.png}
		\vspace{0.2cm}
	\end{subfigure}
    	\hfill
	\begin{subfigure}[b]{0.35\textheight}
		\centering
		\includegraphics[width=\textwidth]{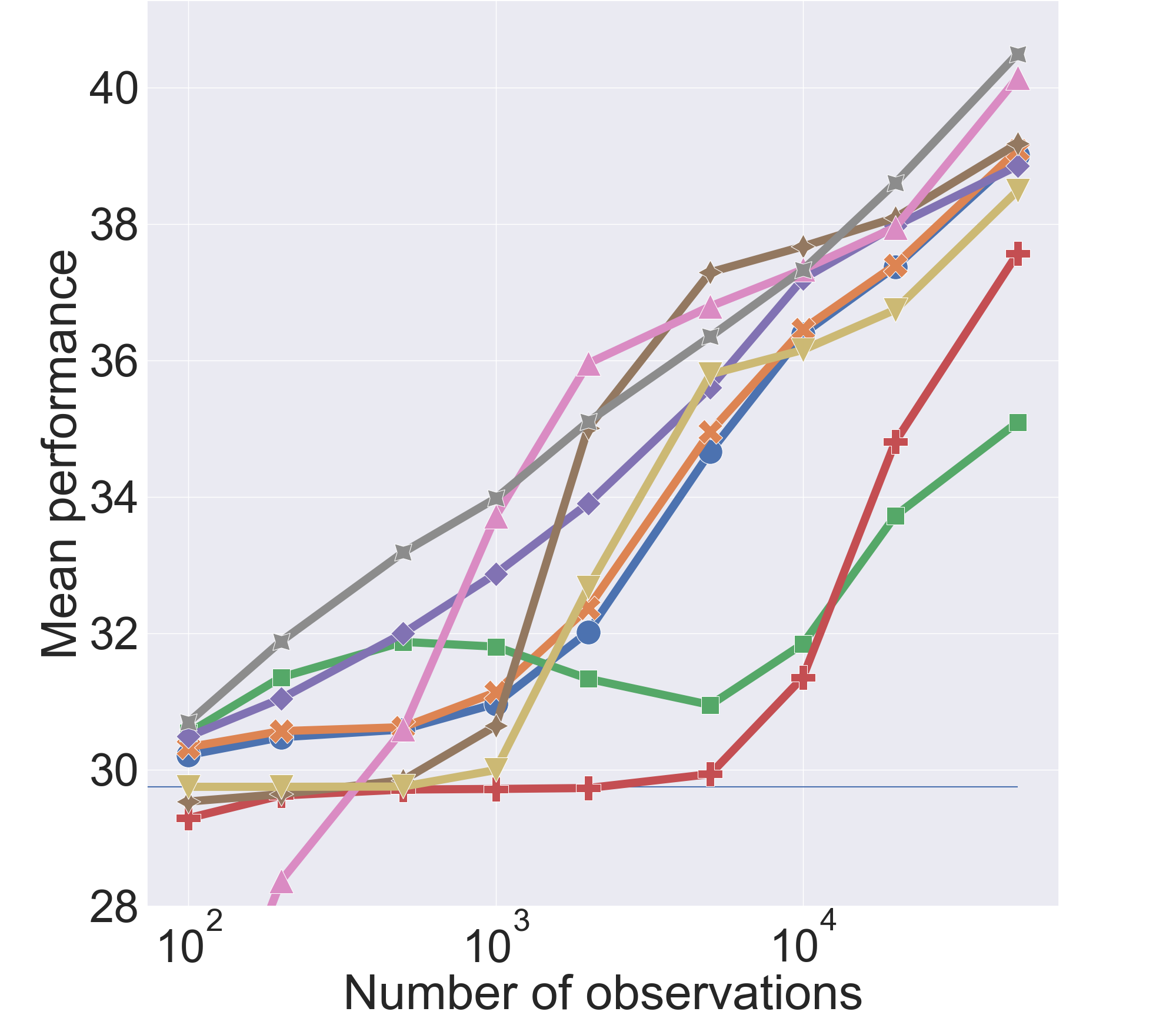}
		\caption{Mean}
		\label{fig:wet-chicken-mean}
	\end{subfigure}
	\hfill
    \begin{subfigure}[b]{0.35\textheight}
		\centering
		\includegraphics[width=\textwidth]{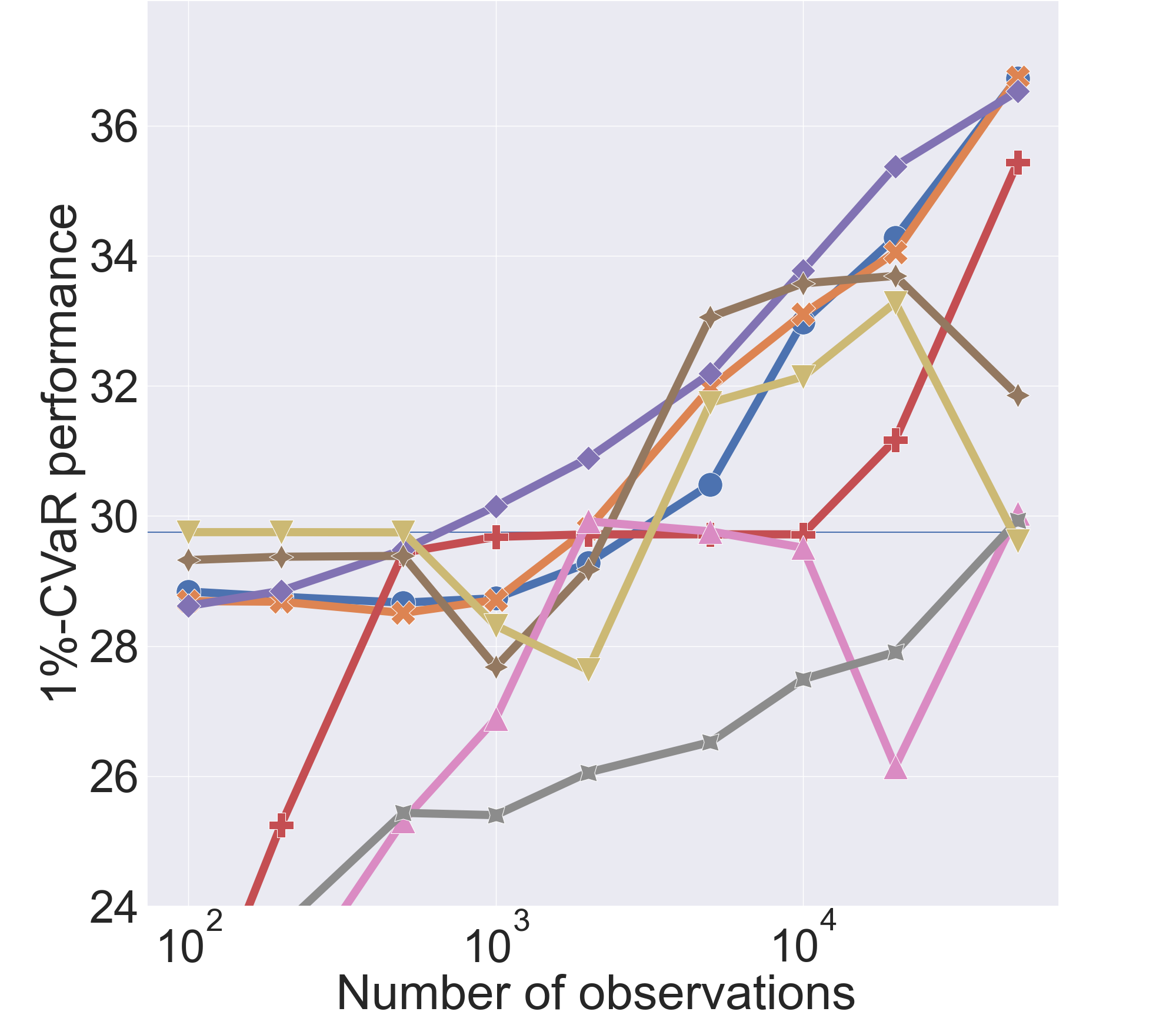}
		\caption{1\%-CVaR}
		\label{fig:wet-chicken-cvar}
	\end{subfigure}
	\caption{Mean (a) and 1\%-CVaR (b) performance over 10,000 trials on the Wet Chicken benchmark for $\epsilon=0.1$ for the baseline policy. The mean performance is dominated by RaMDP, while the restricting algorithms are winning in the risk-sensitive 1\%-CVaR measure. Among the SPIBB class, Lower-Approx-Soft-SPIBB shows the best performance in both runs.}
	\label{fig:wet-chicken}
\end{figure*}
	
	The person has 5 actions, which are ($a_x$ and $a_y$ describe the
	influence of an action on $x_t$ and $y_t$, respectively):
	\begin{itemize}
		\item Drift: The person does nothing, in formula $(a_x,a_y)=(0,0)$.
		\item Hold: The person paddles back with half their power, in formula
		$(a_x,a_y)=(-1,0)$.
		\item Paddle back: The person wholeheartedly paddles back, in formula
		$(a_x,a_y)=(-2,0)$.
		\item Right: The person tries to go to the right parallel to the waterfall, in
		formula $(a_x,a_y)=(0,1)$.
		\item Left: The person tries to go to the left parallel to the waterfall, in
		formula $(a_x,a_y)=(0,-1)$.
	\end{itemize}
	
	The new position of the person assuming no river constraints is then calculated
	by 
	\begin{equation} \label{eq:wet-chicken-rounding}
		(\hat{x},\hat{y}) = (round(x_t + a_x + v_t + \tau_t  s_t),
		round(x_t + a_y))
	\end{equation}
	where the $round$ function is the usual one, i.e., a number is getting rounded
	down if the first decimal is 4 or less and rounded up otherwise. Incorporating
	the boundaries of the river yields the new position as
	\begin{equation}
		x_{t+1}=\begin{dcases}
			\hat{x},& \text{if } 0\leq\hat{x}\leq4 \\
			0, & \text{otherwise}
		\end{dcases}
	\end{equation}
	and
	\begin{equation}
		y_{t+1} = \begin{dcases}
			0,& \text{if } \hat{x}>4 \\
			4,& \text{if } \hat{y}>4 \\
			0,& \text{if } \hat{y}>0 \\
			\hat{y}, & \text{otherwise}
		\end{dcases}.
	\end{equation}
	
	As the aim of this experiment is to have a realistic setting for Batch RL, we use a realistic behavior policy. Thus, we do not incorporate any knowledge about the transition probabilities or the optimal policy as it has been done for the Random MDPs benchmark. Instead we devise heuristically a policy, considering the overall structure of the MDP.
	
	Our behavior policy follows the idea that the most beneficial state might lie in the middle of the river at $(x,y)=(2,2)$. This idea stems from two trade-offs. The first trade-off is between low rewards for a small $x$ and a high risk of falling down for a big $x$ and the second trade-off is between a high turbulence and low velocity for a low $y$ and the opposite for big $y$. To be able to ensure the boat stays at the same place turbulence and velocity should both be limited. 
	
	This idea is enforced through the following procedure. If the boat is not in the state $(2,2)$, the person tries to get there and if they are already there, they use the action \textit{paddle back}. Denote this policy with $\pi_b'$. The problem with this policy is that it is deterministic, i.e., in every state there is only one action which is chosen with probability 1.
	This means that for each state there is at most 1 action for which data is
	available when observing this policy. This is countered
	by making $\pi_b'$ $\epsilon$-greedy, i.e., define the behavior policy $\pi_b$
	as the mixture
	\begin{equation}
		\pi_b = (1-\epsilon)\pi_b' + \epsilon \pi_u
	\end{equation}
	where $\pi_u$ is the uniform policy which chooses every action in every state
	with the same probability. $\epsilon$ was chosen to be 0.1 in the following experiments.

Again, the experiment was run 10,000 times for each algorithm and each hyper-parameter and show the mean and 1\%-CVaR performance in Figure \ref{fig:wet-chicken} for the optimal hyper-parameter, as displayed in Table \ref{tab:hyper-parameter}. Apart from DUIPI, the results are similar to those on the Random MDPs benchmark. Again, the mean performance of R-MIN is extremely bad for few data but then improves strongly. Basic RL
and DUIPI exhibit the worst mean performance. All algorithms from the SPIBB
family perform very well, especially Lower-Approx-Soft-SPIBB, and are only beaten by RaMDP.

Once more, the $1\%$-CVaR performance is of high interest for us and Figure \ref{fig:wet-chicken} confirms many of the observations from the Random MDPs benchmark as well. We find again that especially Basic RL\textemdash{}not even visible in the plot due to its inferior performance\textemdash{}, but also RaMDP and R-MIN have problems
competing with the SPIBB and Soft-SPIBB algorithms. Overall,
Lower-Approx-Soft-SPIBB performs the best, followed by
Adv-Approx-Soft-SPIBB and Approx-Soft-SPIBB. 

These two experiments demonstrate that restricting the set of policies instead of adjusting the action-value function can be very beneficial for the safety aspect of RL, especially in complex environments and for a low number of observations. On the contrary, from a pure mean performance point of view it is favorable to rather adjust the action-value function. 

\section{Conclusion}
\label{sec:conclusion}
We show that the algorithms proposed in \cite{nadjahi_safe_2019} are not provably safe and propose a new version that is provably safe. We also adapt their ideas to derive a heuristic algorithm which shows, among the entire SPIBB class on two different benchmarks, both the best mean performance and the best $1\%$-CVaR performance, which is important for safety-critical applications.
Furthermore, it proves to be competitive in the mean performance against other state of the art uncertainty incorporating algorithms and especially to outperform them in the $1\%$-CVaR performance. Additionally, it has been shown that the theoretically supported Adv-Approx-Soft-SPIBB performs almost as well as its predecessor
Approx-Soft-SPIBB, only falling slightly behind in the mean performance. 

The experiments also demonstrate different properties of the two classes of SPI algorithms in Figure~\ref{fig:taxonomy}: algorithms penalizing the action-value functions tend to perform better in the mean, but lack in the 1\%-CVaR, especially if the available data is scarce.

Perhaps the most relevant direction of future work is 
how to apply this framework to continuous 
MDPs, which has so far been explored by \cite{nadjahi_safe_2019} without theoretical safety guarantees. Apart from theory, we hope that our observations of the 
two classes of SPI algorithms can contribute to the 
choice of algorithms for the continuous case.

\bibliographystyle{unsrtnat}
\bibliography{main}  






\end{document}